\documentclass{article}
\pdfoutput=1

\usepackage{arxiv}
\usepackage[utf8]{inputenc} 
\usepackage[T1]{fontenc}    
\usepackage{hyperref}       
\usepackage{url}            
\usepackage{booktabs}       
\usepackage{amsfonts}       
\usepackage{nicefrac}       
\usepackage{microtype}      
\usepackage{lipsum}		
\usepackage{graphicx}
\usepackage{natbib}
\usepackage{amssymb}
\usepackage{amsmath,bm}
\usepackage{multirow}
\usepackage{adjustbox}
\usepackage{amsfonts}
\usepackage[ruled,vlined]{algorithm2e}
\usepackage{caption}
\usepackage{subcaption}

\title{Graph-based Joint Pandemic Concern and Relation Extraction on Twitter}
\date{}

\author{Jingli Shi \\
	Engineering, Computer and Mathematical Sciences \\
	Auckland University of Technology \\
	Auckland, New Zealand \\
	\texttt{jingli.shi@aut.ac.nz} \\
	\And
	Weihua Li \\
	Engineering, Computer and Mathematical Sciences \\
	Auckland University of Technology \\
	Auckland, New Zealand \\
	\texttt{weihua.li@aut.ac.nz} \\
	\And
	Sira Yongchareon \\
	Engineering, Computer and Mathematical Sciences\\
	Auckland University of Technology\\
	Auckland, New Zealand \\
	\texttt{sira.yongchareon@aut.ac.nz} \\
	\And
	Yi Yang \\
	Applied Artificial Intelligence Institute\\
	Deakin University\\
	Victoria, Australia \\
	\texttt{y.yang@deakin.edu.au} \\
	\And
	Quan Bai \\
	Information and Communication Technology\\
	University of Tasmania\\
	Hobart, Australia \\
	\texttt{quan.bai@utas.edu.au} \\
}

\renewcommand{\shorttitle}{\textit{arXiv} Template}

\begin{document}
\maketitle

\begin{abstract}
Public concern detection provides potential guidance to the authorities for crisis management before or during a pandemic outbreak. Detecting people's concerns and attentions from online social media platforms has been widely acknowledged as an effective approach to relieve public panic and prevent a social crisis. However, detecting concerns in-time from massive information in social media turns out to be a big challenge, especially when sufficient manually labelled data is in the absence of public health emergencies, e.g., COVID-19. In this paper, we propose a novel end-to-end deep learning model to identify people's concerns and the corresponding relations based on Graph Convolutional Network and Bi-directional Long Short Term Memory integrated with Concern Graph. Except for the sequential features from BERT embeddings, the regional features of tweets can be extracted by the Concern Graph module, which not only benefits the concern detection but also enables our model to be high noise-tolerant. Thus, our model can address the issue of insufficient manually labelled data.  We conduct extensive experiments to evaluate the proposed model by using both manually labelled tweets and automatically labelled tweets. The experimental results show that our model can outperform the state-of-art models on real-world datasets. 
\end{abstract}



\section{Introduction}
\label{intro}
The outbreak of coronavirus (COVID-19) in 2019 has been causing a rapid increase in both infection and death rates around the world. Especially when the pandemic moved into the second wave, it caused devastating loss of human lives, impacted the global economy, transformed our daily lives, and posed a threat to our society (\cite{killgore2020loneliness}). According to the studies on the past pandemic outbreaks, e.g., Zika, Ebola, and H1N1, social media platforms, e.g., Twitter,  have proven to be a popular channel for spreading information, especially related to public opinions and concerns (\cite{damiano2020content}). This is because people intend to perceive more details regarding the pandemic by reading the newsfeeds and interpreting the comments from others through social networks (\cite{li2018automated,hu2019context}).  Twitter, a popular and informative social network platform, allows people to post and interact with messages known as ``tweets". They can also communicate and express opinions about the latest events (\cite{killgore2020loneliness}). User-generated tweets from Twitter turn out to be ``prophetic", namely, valuable indicators of what issues will likely happen in the pandemic. It is important to make use of tweets and investigate what various people are discussing during the pandemic.  The attitudes and behaviours of our society are affected directly by public concerns. Thus, how to effectively extract public concerns and analyse the corresponding relationships will assist people in understanding the anxiety and fears of the society in this pandemic situation. Furthermore, the potential social crisis also can be revealed by analysing public concerns, which significantly contributes to social management control.

Motivated by this background, a great many efforts have been dedicated to mining social media data and exploring the opinions towards pandemic outbreaks (\cite{da2020publishing}). Most existing research works can be categorised into traditional survey methods, i.e., survey and questionnaire (\cite{nelson21rapid} and machine learning model-based methods, i.e., topic modelling (\cite{van2020women, kassab2020nonnegative}). The existing studies are capable of extracting fundamental public concerns, e.g., ``social distancing",  ``hand sanitizer" and  ``face masks", which require intensive human-effort in labelling large datasets, turning out to be inefficient. Moreover, in any epidemic emergence situation, e.g., COVID-19, traditional approaches, such as questionnaires and clinical tests, neither collect enough data for deep learning model training nor rapidly generate a model for concern detection. Therefore, it is vital to design an end-to-end model that is capable of automatically analysing social media data and detecting public concerns without requiring a large-scale of data to be labelled manually.

Deep learning methods are increasingly applied to valuable information extraction. However, most methods highly rely on data labelled by the annotators, requiring much time and financial resources (\cite{kipf2017semi}). Moreover, the noisy and imbalanced social media data prevent deep learning-based methods from generalisation (\cite{rathan2018consumer}). In many existing studies, the proposed models are not able to track real-time statistics of public concerns related to pandemics due to the required labelled dataset (\cite{li2020characterizing, jahanbin2020using, hou2020assessment, lazard2015detecting}). To mitigate this issue, we have conducted preliminary research to mine public concerns by proposing an Automated Concern Exploration (ACE) framework (\cite{shi2020concern}). The proposed framework can detect concerns from tweets automatically and construct a concern knowledge graph to present the interconnections of the extracted concern entity set. However, several advent limitations are still to be addressed. (1) only BERT embedding of tweets is used, which cannot capture regional dependency word features from tweets to improve the performance of concern extraction. (2) the relation between concerns in one tweet posted by a user is not detected, which is critical to reveal meaningful information about public concerns. (3) The framework employs a rule-based method, having poor generalisability and appearing difficult to transfer to future occurring pandemics.

In this paper, we propose and develop an end-to-end model with concern graph and concern states to identify public concerns and corresponding relations simultaneously. We formally define ``public concern" with a consideration of its type and degree, and construct concern graph to represent the regional features, improving the concern identification effectiveness. Furthermore, by integrating concern states with Graph Convolutional Network (GCN)  (\cite{kipf2017semi}, the proposed method can extract concern relations. Extensive experiments are conducted to evaluate the proposed method by using both manual-labelled and auto-labelled datasets. The experimental results explicitly demonstrate that our method outperforms state-of-the-art models.

Our contributions in this research work are summarized below:
\begin{itemize}
	\item We define a concern graph data structure to capture the inherent structural information of concerns more efficiently.
	\item We present a novel end-to-end model to jointly extract concerns and relations consisting of CG and shared state of concerns.
	\item We evaluate our model on manual-labelled data and auto-labelled data, and the results indicate the proposed method is effective for auto-labelled data.
\end{itemize}

\section{Related Work}
\label{sec:1}

In this section, we first review the existing studies related to public concern mining and detection. Then, modern approaches of Named Entity Recognition (NER) and Relation Extraction (RE) are inspected and compared since the concern detection, defined in this paper, tends to explore the concern entities and the corresponding relations.  Finally, we review the GCN and its variants since GCN has been widely adopted in NER and RE based on recent studies.

\subsection{Concern Detection}
Social media has become a prevalent platform for people to communicate and express their opinions. With the outbreaks of the corona-virus pandemic, i.e., COVID-19, how to effectively extract people's opinions and address the public concern in pandemic situations has attracted great attention to researchers. Thus, great efforts have been dedicated to the analysis of public response to pandemics in social media platforms, e.g., Twitter. The current approaches are mainly categorized into two types of methods: probabilistic model-based and deep learning-based. In probabilistic-based models, Latent Dirichlet Allocation (LDA) is commonly used for public concern extractions. For example,  Chandrasekaran et al. conduct a temporal assessment on COVID-19-related tweets, aiming to uncover public concern trends through extracting topics and predicting sentiment scores (\cite{chandrasekaran2020topics}). Xue et al. utilise LDA to analyse public response towards COVID-19 pandemic on the social media platform, which aims to identify popular uni-grams and bi-grams topics from tweets (\cite{xue2020twitter}). Wahbeh et al. adopt qualitative analysis tool to detect recommendations, topics, and opinions related to COVID-19 pandemic from Twitter (\cite{wahbeh2020mining}). Whereas, probabilistic model-based methods perform poorly on public concern identification since contextual information is ignored. By contrast, deep learning-based methods are able to retain contextual features of sentences. Nowadays, deep learning is widely adopted as a popular approach for many NLP tasks, e.g., sentiment analysis. By employing such an approach, many studies aim to extract insightful information for assisting the authorities in making appropriate responses and reactions (\cite{wang2020covid,yin2020detecting,chen2020eyes}). However, most existing research works only identify a few pre-defined public concerns but neglect the relations between the concerns. Without concern relations, it is difficult to identify the cause of public concerns or reveal people's thoughts behind the expressed concerns. Different from the above two types of approaches, our proposed method is able to capture regional and sequential features of a sentence and assist the extraction of public concerns with the corresponding relations. 

\subsection{Named Entity Recognition}
\label{sec:2}
NER, also named entity extraction, is one of the classic tasks of Natural Language Processing (NLP), which aims to identify the classify entities from unstructured text into pre-defined categories (\cite{mohit2014named}). Recent studies have shown two typical NER approaches, i.e., traditional statistical models and deep learning-based methods. Zhou et al. propose an entity extraction model with a chunk tagger method based on Hidden Markov Model (HMM), and the model outperforms the hand-crafted rules-based models (\cite{zhou2002named}). Lafferty et al.  present the Conditional Random Fields (CRF), to segment and label sequence data by building a probabilistic model (\cite{lafferty2001conditional}). However, traditional statistical models perform poorly on complex sentences  because they fail to discover hidden features from data. Compared with traditional methods, deep learning-based approaches are able to learn latent representations from raw data and achieve promising performance.  By relying on character and word representations, a novel neural architecture is introduced, which combines bidirectional Long Short-Term Memory (BiLSTM) and CRF (\cite{lample2016neural}). Ma et al. propose a novel deep learning-based model by combining biLSTM, Convolutional Neural Network (CNN), and CRF (\cite{ma2016end}). Recently, the current SoTA models adopt context-dependent embeddings, e.g., ELMo ((\cite{peters2018deep}), Flair ((\cite{akbik2018contextual}), and BERT ((\cite{devlin2018bert}), to encode the input. Although deep learning-based models are capable of capturing contextual features of data, interaction information between entities is neglected. Different from the above models, apart from contextual information, we also propose a designated Concern Graph (CG) to capture specific features of entities, enabling our method to perform better on Twitter data.

\subsection{Relation Extraction}

As a fundamental task in the NLP field, Relation extraction (RE) aims to detect and classify the semantic relationship between entity mentions (\cite{chinchor1998overview}). Early research works mainly focus on rule-based models, in which proper rules are difficult to define without domain knowledge. To address such an issue, many efforts have been dedicated to kernel-based models with manual-labelled data (\cite{culotta2004dependency, zhou2011kernel, seewald2007lambda}). The key weakness of kernel-based methods is that contextual features are not captured, leading to wrong relation extraction on data with a long sentence. Recently, deep neural networks are applied to relation extraction due to their supremacy in terms of accuracy. Therefore, some popular deep learning models, e.g., CNN, LSTM, GCN, are utilised to learn contextual features of data and achieve better performance than kernel-based models (\cite{zeng2014relation, miwa2016end, fu2019graphrel}). 

Apart from the models extracting entity and relation separately, many other research works exploring joint methods to extract both simultaneously. Arzoo et al.  propose an attention-based RNN model for joint entity mentions and relations extraction (\cite{katiyar2017going}). Zheng et al.  present a novel tagging strategy to covert sequence labelling and classification tasks to a tagging problem and extract entities and relations directly using the joint model (\cite{zheng2017joint}). Miwa et al.  use Tree-LSTM with bidirectional sequential LSTM to extract entity and relation simultaneously (\cite{miwa2016end}).  Zeng et al. propose a sequence-to-sequence model with a copy mechanism to extract entity and relation (\cite{zeng2018extracting}). However, existing NER, RE, and  joint entity and relation extraction models suffer from two issues. First, existing models only discover contextual features of a sentence and neglect entity features, which is vital for entity extraction. Second, relation extraction mainly rely on contextual features of a sentence and the information of entities corresponding to a relation is ignored. This can be a severe problem for social media data, where numerous grammatical mistakes exist in sentences. To address these two issues, we combine contextual and concern features for concern identification and integrate learned concern features with the module of relation extraction.

\subsection{Graph Convolutional Network}
Graph Convolutional Network (GCN) has demonstrated advent advantages in capturing the dependency structure of sentences, and it has been widely adopted in many NLP tasks (\cite{battaglia2016interaction, defferrard2016convolutional, hamilton2017inductive}). Hong et al.  present a joint model based on GCN to perform entity and relation extraction by considering  context and syntactic information of sentences  (\cite{hong2020joint}).  Zhang et al.  utilise GCN over a pruned dependency tree to tackle the relation extraction task (\cite{zhang2018graph}).  Inspired by the existing studies, we incorporate GCN into the proposed model to effectively preserve the dependency information of sentences. Furthermore, concern states are integrated with GCN to improve the accuracy of relation extraction.

In this paper, we proposed an end-to-end model with a concern graph module to perform joint extraction of concerns and relations. Meanwhile, we integrate the concern states from BiLSTM with the input features of BiGCN to enhance the influences from concerns to improve relation extraction performance.

\begin{figure}[h]
\centering
  \includegraphics[width=\textwidth]{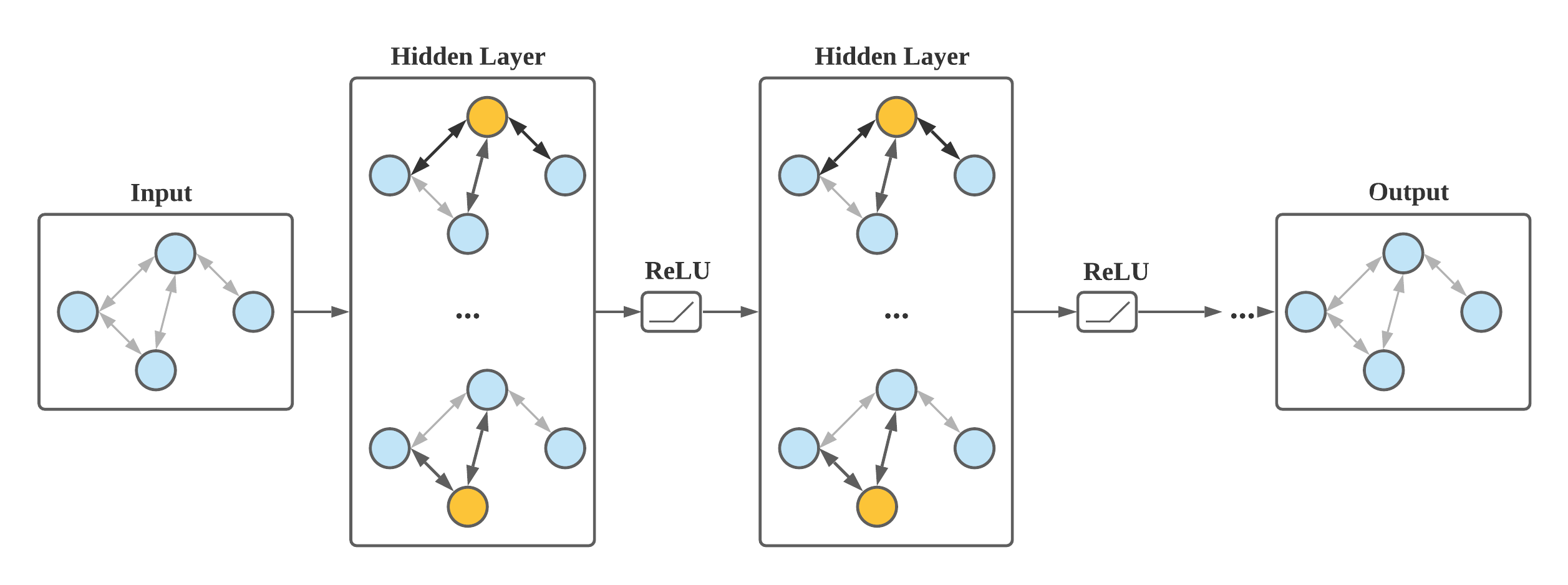}
\caption{Bidirectional Graph Convolutional Network (BiGCN)  Overview}
\label{fig:bigcn}       
\end{figure}

\section{Preliminaries}
In this section, the relevant definitions are presented, including public concerns, concern relations, and graph. In addition, the concern detection problem is formally formulated. 
\subsection{Formal Definition}

\textbf{Definition 1: Concern} refers to people worry a real or imagined issue. Public concern represents a word or a phrase in a tweet towards which most people express strong opinions about a particular aspect of the pandemic. Given a concern set $C=\{c_1,..., c_n\}$, the $i$th potential concern detected in tweet $t_j$ can be defined as $c_i^j = (ce_i^j, cs_i^j)$, where $ce_i^j \in CE$ is the concern entity identified in tweet $t_j$ and it can be words or phrase, e.g., \emph{``China"},  \emph{``corona emergency relief"} and \emph{``florida medical examiner"}, and $cs_i^j$ is concern score of concern $c_i^j$ and the value  is calculated by use of Equation \ref{equation:cs}, in which $rt_i^j$ represents the retweet count of tweet $t_j$. $sp_i^j \in [-1, 1]$ denotes the sentiment polarity of tweet $t_j$, where -1 indicates an extremely negative attitude, 0 means a neutral attitude and +1 implies an extremely positive attitude. The range of $cs_i^j$ is [0,1], where the greater the value is, the more likely it becomes a concern.  $C = \{c_i^t | i \in [1,N], t \in T\}$ denotes the set of public concerns detected in Twitter dataset $T$. For each concern $c_i$, there is one attribute named type $ct_i$, where $ct_i \in CT$ and $CT = \{ct_1, ..., ct_n  \}$ is the set of concern types. 

\begin{equation}
\label{equation:cs}
cs^j = (1- \theta) * |sp^j| + \theta * {\tilde{rt}}^j ,
\end{equation}
where $\theta \in [0, 1]$ refers to the weight parameter and $\tilde{rt}^j$ describes the normalized value of $rt^j$.

\textbf{Definition 2: Concern Relation} describes the relationship between public concern pair. We use $r_{m,n}^j \in R$ to present the relation between concern $c_m^j$ and $c_n^j$ in tweet $t_j$, where $r_{m,n}^j$ is unidirectional relation, i.e., the same as $r_{n,m}^j$, and $R$ is the set of relation extracted from Twitter dataset $T$.

\textbf{Definition 3: Concern Triple} is the fundamental element of the public concern graph which is extracted from a tweet. Because some short words or phrases are very limited in context information to present the real meaning of concern. Whereas, the concern triple is capable of semantically representing what a concern is about. A public concern triple in the tweet $t_j$, $ct_{m,n}^j = (s_m^j, r_{m,n}^j, o_n^j)$, has three components, i.e., $s_m^j$, $r_{m,n}^j$ and $o_n^j$, referring to as the subject, relation, and object of the concern triple, respectively. The $s_m^j$ and $o_n^j$ are extracted entities and $r_{m,n}^j$ is the extracted relation based on dependency parser analysis of the tweet $t_j$.

\begin{figure}[h]
\centering
  \includegraphics[width=\textwidth]{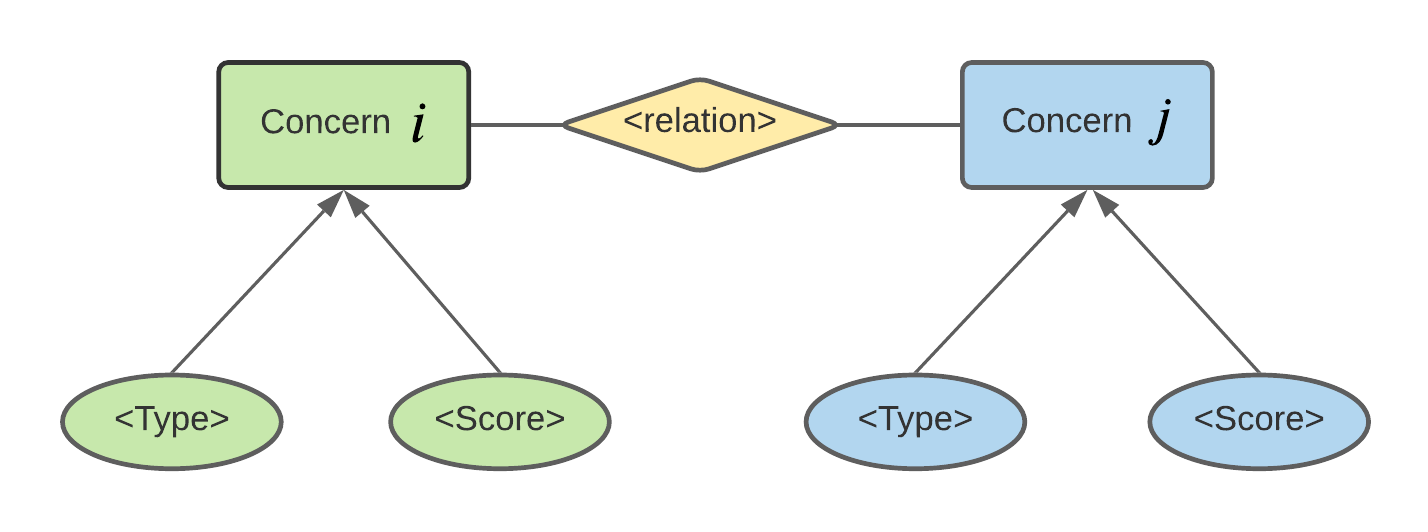}
\caption{Concern Graph (CG): each concern has two attributes, i.e., type and score, along with relation to another concern to form the concern graph. }
\label{fig:concern_graph}     
\end{figure}

\textbf{Definition 4: Concern Graph (CG)} aims to explore discriminative public concerns and what kind of relations are existing between concerns. In order to present the relation of public concerns, Concern Graph (CG) is proposed as the control signal for capturing public concerns. CG of the tweet $t_j$ can be denoted as $G = (\nu, \varepsilon)$, where $\nu$ is the set of nodes, and $\varepsilon$ is the edges set. As shown in Figure \ref{fig:concern_graph}, nodes of CG are classified into four categories:  (1) object $o^j$, subject node $s^j$; (2) relation note $r^j$; (3) attribute node $a^j$ including concern type $ct^j$; (4) concern score $cs^j$. 

The CG $G$ is constructed via the following steps:   

\begin{enumerate}
	\item Detect public concern $c_i^j$ and add it to $G$, where $c_i^j$ is grounded in the tweet $t_j$.
	\item Extract the descriptive details of concern $c_i^j$ as the attribute $a_{i,l}^j$ including type $ct_i^j$ and score $cs_i^j$, then add them to $G$ and assign an un-directed edge from $c_i^j$ to $a_{il}^j$, where $|l|$ is the number of attributes towards concern $c_i^j$.
	\item Identify the relation $r_{ik}$ between concerns $c_i^j$ (subject in concern triple) and $c_k^j$ (object in concern triple), which is a unidirectional type of relation. Adding relation node $r_{ik}$ to $G$ and assigning edges from $c_i^j$ to $r_{ik}$ and from $r_{ik}$ to $c_k^j$.
\end{enumerate}

\subsection{Problem Formulation}
In the previous section, we describe the definitions related to our proposed method. Based on the definitions, our model aims to jointly extract typical concerns $\{c_{i}^{j} | c_{i}^{j} \in C \wedge p_{i}^{j} < 0 \}$ from tweet $t_j$ and concern relations $r_{mn}$ between concern $c_m$ and $c_n$ from Twitter dataset $T$ by constructing Concern Graph (CG).

\section{Graph-based Concern and Relation Extraction} 
\begin{figure}
\centering
  \includegraphics[width=1.0\textwidth]{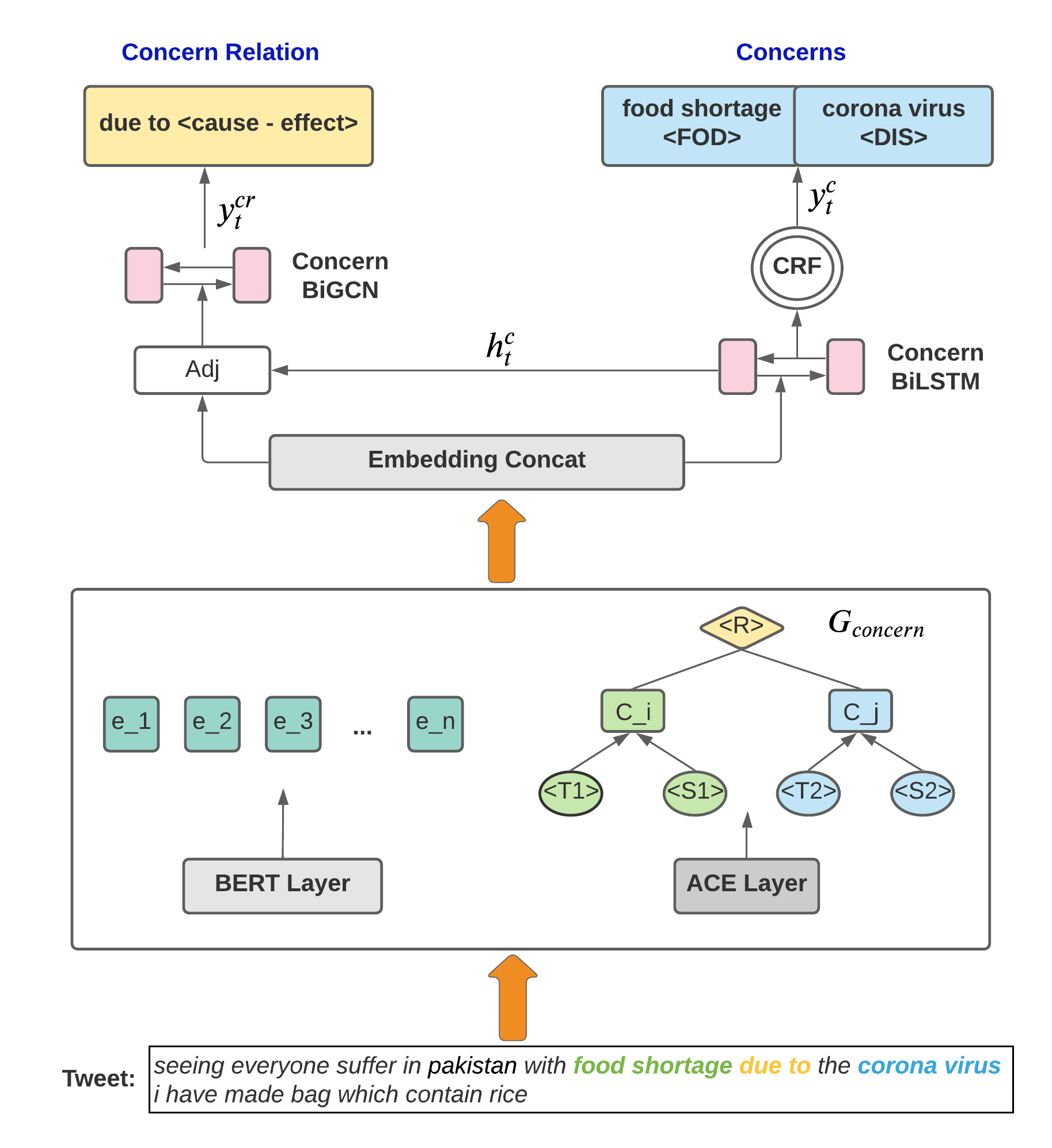}
\caption{The overview of CG-CRE model. }
\label{fig:proposed_model}    
\end{figure}

For a tweet $T$, the goal of our method is to identify public concerns $C=\{c_1, ..., c_n\}$ and concern relations $R=\{r_1, ..., r_n\}$. In this section, the joint extraction of concerns and relations model with concern graph is illustrated in Figure \ref{fig:proposed_model}. The proposed method consists of four main components, i.e., Concern and Relation Extraction (CRE) embedding layer, Concern and Relation Extraction (CRE) encoding layer, concern decoding layer, and concern relation extraction layer. We describe each component in detail below. Concern and Relation Extraction (CRE) Embedding layer is introduced in Section \ref{cre embedding layer}, followed by Concern and Relation Extraction (CRE) encoding in Section \ref{cre encoding layer}. Concern decoding and concern relation extraction layer are presented in Sections \ref{concern decoding layer} and \ref{relation extraction layer}, respectively. The model objective function is explained in Section \ref{model objective function}.

\begin{algorithm}[H]
\caption{Training Process of CG-CRE Model }
\SetAlgoLined
	\SetKwInOut{Input}{Input}
    \SetKwInOut{Output}{Output}
 	\Input{ $T$, $\upsilon_w$, $\upsilon_{cg}$, $E$, $B$, $lr$, $d$ \\

			$T$ indicates labelled Twitter corpus for model training  \\
			$\upsilon_w$ represents BERT embedding of Twitter corpus  \\
			$\upsilon_{cg}$ is embedding of CG  \\
			$E$ is epoch number \\
			$B$ is batch size \\
			$lr$ indicates learning rate  \\
			$d$ is embedding dimension \\
			 }
	\Output{
				$L$  \\
				$L$ is the loss function value  \\
			}

 initialization: wight vector $W$\;
 \While{$ep$ in $EP$}
{
	\While{$b$ in $B$}
	{

			generate embedding  $\grave{X^{(b)}}$, $\acute{X}^{(b)}$\;
			compute concern hidden state ${h_t}'^{(b)}$ as Equation \ref{concern_hidden} \;
			$L_{(c)}^{(b)} = max( \sum log(P_{(c)}^{(b)}=S_{(c)}^{(b)} ) ) $ \;
			$L_{(r)}^{(b)} =  max( \sum log(P_{(r)}^{(b)}=S_{(r)}^{(b)} ) ) $ \;
			$L^{(b)} =  L_{(c)}^{(b)} + \alpha * L_{(r)}^{(b)} $ \;
	}
 }
 
\end{algorithm}

\subsection{Concern and Relation Extraction (CRE) Embedding Layer}
\label{cre embedding layer}
Since deep learning models are integrated into our method, word tokens and proposed CG need to be transformed into low-dimensional vectors by embedding layer. For a tweet $t = \{w_1, ...,w_i, ..., w_n\}$, where $w_i$ denotes the $i$th word in the tweet. Given the tweet $t$, pre-trained BERT model is used to generate word embedding set $\grave{X} = \{ \tilde{e_1}, ... \, \tilde{e_i}, \, ... , \tilde{e_n} \, |\, \tilde{e_i} \in \mathbb{R}^d \}$, where $\tilde{e_i}$ represents the embedding of word $w_i$ and $d$ means the embedding dimension. 

In order to enhance model input features, we further encode proposed Concern Graph (CG) $G$ to obtain CG node embedding $\acute{x}_i^{(0)}$ as below:

\begin{equation}
\centering
\label{cg_embedding}
\acute{x}_i^{(0)}=\left\{\begin{matrix}
(v_i^{(dep)} + v_i^{(pos)}) \, \odot \, W_{cr}[0] , & if \, i \in C;\\ 
v_i^a \, \odot \, W_{cr}[1], &  if\,  i \in A ;\\ 
v_i^r \, \odot \, W_{cr}[2], & if\,  i \in R; 
\end{matrix}\right.
\end{equation}
where $v_i^{(dep)}$ and $v_i^{(pos)}$ denote the syntactic dependency relation and POS tag feature, respectively. Both  $v_i^{(dep)}$ and $v_i^{(pos)}$ are used to capture the meaning of tweet and words syntactic dependency. $C$ represents the concern set. $v_i^a$ represents the attribute features including concern type and score. $A$ means attribute set. $v_i^r$ indicates relation feature, and $R$ is relation set. $ W(\cdot ) \in \mathbb{R}^{3\times d} $ refers to parameters, where $d$ means the feature dimension.

\subsection{Concern and Relation Extraction (CRE) Encoding Layer}
\label{cre encoding layer}
In order to capture long-distance dependencies and forward and backward features between tokens in tweets, bidirectional LSTM is used in this paper. The Bi-LSTM contains forward and backward layers, and a concatenation layer of backward and forward state information. The embeddings in Section \ref{cre embedding layer} are concatenated as the input of the concern encoder layer.  
The BiLSTM encoding layer is defined by using Equations \ref{lstm_i} - \ref{concern_hidden}:

\begin{equation}
\label{lstm_i}
i_t = \sigma (W_{ex}^{(i)}*[\tilde{e_i};\acute{x}_j^{(i)}] + W_h^{(i)}*h_{t-1}+b^{(i)})
\end{equation}
\begin{equation}
\label{lstm_f}
f_t = \sigma(W_{ex}^{(f)}*[\tilde{e_i};\acute{x}_j^{(f)}] + W_h^{(f)}*h_{t-1}+b^{(f)})
\end{equation}
\begin{equation}
\label{lstm_o}
o_t = \sigma(W_{ex}^{(o)}*[\tilde{e_i};\acute{x}_j^{(o)}] + W_h^{(o)}*h_{t-1}+b^{(o)}) 
\end{equation}
\begin{equation}
\label{lstm_u}
u_t = \sigma(W_{ex}^{(u)}*[\tilde{e_i};\acute{x}_j^{(u)}] + W_h^{(u)}*h_{t-1}+b^{(u)})
\end{equation}
\begin{equation}
\label{lstm_c}
c_t = i_t \odot u_t + f_t \odot c_{t-1}
\end{equation}
\begin{equation}
\label{concern_hidden}
h_t = o_t \odot tanh(c_t) , 
\end{equation}
where $\sigma$ is sigmoid activation function, $W(\cdot)$ refers to weight parameters. and $[;]$ is a vector concatenation operation. $\tilde{e_i}$ and $\acute{x}_j(\cdot)$ denote word embedding and embedding of CG $G$ defined in Section \ref{cre embedding layer}. In Equations \ref{lstm_i} - \ref{lstm_u}, $b(\cdot)$ refers to the bias vector, and $\odot$ represents element-wise multiplication. $c$ and $h$ denote cell state and hidden state, respectively, carrying information from the previous layer to the next layer. Because Bi-LSTM is applied in our method, the hidden state is obtained by concatenating both direction hidden state, namely, forward direction $\overrightarrow{{h_t}'}$ and backward direction $\overleftarrow{{h_t}'}$, therefore, the final hidden state can be denoted as ${h_t}' = [\overrightarrow{{h_t}'}, \overleftarrow{{h_t}'}]$. By passing hidden state to a fully connected neural network, the final output of Bi-LSTM can be defined as:

\begin{equation}
\centering
O = W^o * {h_t}' + b^o ,
\end{equation}
where $W^o$ is the output weight parameters and $b^o$ is the bias vector.

\subsection {Concern Decoding Layer}
\label{concern decoding layer}

In our model, the CRF is employed to produce a tag sequence since it can produce a higher tagging accuracy than that of the existing models 
(\cite{hong2020joint}). For one tweet $t=\{ w_1, ..., w_n \}$, we aim to predict the concern tag sequence $Y^{(c)} = \{y^{(c)}_1, y^{(c)}_2, ..., y^{(c)}_n \}$ where $n$ denotes the number of words and superscript $(c)$ means the notation of concern. Thus, the CRF score can be defined as in Equation \ref{eq_prediction_score}:

\begin{equation}
\centering
\label{eq_prediction_score}
S^{(c)}(t,Y^{(c)}) =  \sum_{i=1}^{n}O_{i,y^{(c)}_i} + \sum_{i=1}^{n}T_{y^{(c)}_i, y^{(c)}_{i+1}} ,
\end{equation}
where $O \in \mathbb{R} ^{n \times k}$ indicates the matrix of scores output from the previous encoding layer with $k$ as the number of distinct tags, and $O_{i,j}$ denotes the score of the $jth$ tag of the $ith$ word in tweet $t$. $T$ represents a matrix of transition scores as being introduced in (\cite{huang2015bidirectional}, and $T_{i,j}$ means the score of a transition from tag $i$ to tag $j$. Then, for input tweet $t$, the probability of a given sequence of tags over the sequence of predicted tags $Y^{(c)}$ is defined by applying the softmax layer  as in Equation  \ref{eq_pd_ct}.

\begin{equation}
\centering
\label{eq_pd_ct}
P^{(c)}= \frac{e^{S^{(c)}(t,Y^{(c)}) }}{\sum_{\widetilde{y}^{(c)} \in Y^{(c)}_X}e^{S^{(c)}(t,\widetilde{y}^{(c)}) }}, 
\end{equation}
In Equation  \ref{eq_pd_ct}, $ Y^{(c)}_X$ denotes all possible concern tag sequences for tweet $t$. 

\subsection {Concern Relation Extraction Layer}
\label{relation extraction layer}

Given concern set $C=\{c_1, ..., c_m\}$ in tweet $t=\{t_1, ..., t_n\}$, we aim to extract corresponding relation $r_i \in R$. Except for sequential features, BiGCN is utilised to capture regional features from the tweets. Both forward and backward directions are considered and the hidden state of BiGCN is defined using Equations \label{eq_bigcn_hf} - \label{eq_bigcn_h}).

\begin{equation}
\label{eq_bigcn_hf}
\overrightarrow{{h_{t}}''} = \varsigma (\sum_{v \in \overrightarrow{N(w)}}(\overrightarrow{W_h}*[\overrightarrow{{h^v_{t-1}}''};\overrightarrow{{h_{t-1}}'}] + \overrightarrow{b}))
\end{equation}
\begin{equation}
\label{eq_bigcn_hf}
\overleftarrow{{h_{t}}''} = \varsigma (\sum_{v \in \overleftarrow{N(w)}}(\overleftarrow{W_h}*[\overleftarrow{{h^v_{t-1}}''};\overleftarrow{{h_{t-1}}'}] + \overleftarrow{b})) 
\end{equation}
\begin{equation}
\label{eq_bigcn_h}
{h_{t}}'' = [\overrightarrow{{h_{t}}}'' \, ;\,  \overleftarrow{{h_{t}}}''],
\end{equation}
where $\varsigma$ represents ReLU activation function, ${h_{t}}'' $ refers to the hidden state at $t$th layer and $\overrightarrow{{h_{t-1}}'}$ indicates the shared hidden state from concern detection module. $\overrightarrow{N(w)}$ describes the neighbours of word $w$ in the forward direction and  $\overleftarrow{N(w)}$ means the neighbours of word $w$ in the backward direction. $\overrightarrow{W_h}$ and $\overleftarrow{W_h}$ represent weight parameters in the forward and backward direction, respectively.  $\overrightarrow{b}$ and $\overleftarrow{b}$  are the bias of the model.  $h^{t}$ refers to the final hidden state of word $w$, concatenating hidden states in both directions.

By using hidden states of BiGCN, the relation tendency score $S^{(r)}_{( r_{ij} |c_i, c_j)}$ is defined in Equation \ref{eq_r_s}.

\begin{equation}
\label{eq_r_s}
S^{(r)}_{( r_{ij} |c_i, c_j)} = W^{(r)}*  \varsigma( W^{(r)}_{c_i} * h''_{c_i} + W^{(r)}_{c_j} * h''_{c_j} + b^{(r)}  ), 
\end{equation}
where, superscript $(r)$ means the notation of concern relation.  $S^{(r)}_{( r_{ij} |c_i, c_j)}$ represents the tendency score of concern relation on concerns pair $(c_i, c_j )$. 
$ W^{(r)}$, $W^{(r)}_{c_i}$ and $ W^{(r)}_{c_j}$ are weight parameters.  $b^{(r)}$ denotes the bias term. We apply the activation function (softmax) to the tendency score  $S^{(r)}_{( r_{ij} |c_i, c_j)}$ to obtain the probability of relation $r_{i,j}$ in Equation \ref{eq_rc_p}).

\begin{equation}
\label{eq_rc_p}
P^{(r)}=  \sigma(S^{(r)}_{( r_{ij} |c_i, c_j)})
\end{equation}

\subsection{Model Objective Function}
\label{model objective function}
In this subsection, the final objective function for model training is described. To train proposed model, we use maximum log-likelihood as the loss function and maximize combined loss functions of concern and relation by using Equations \ref{eqn:global_objective_c}, \ref{eqn:global_objective_r} and \ref{eqn:global_objective}.

\begin{equation}
\label{eqn:global_objective_c}
L_{(c)} = max( \sum_{i=1}^{|\mathbb{R}_T|} \sum_{w=1}^{|W_i|} log(P_w^{(c)}=S_w^{(c)}|t_i,\Theta ) )
\end{equation}

\begin{equation}
\label{eqn:global_objective_r}
L_{(r)} =  max( \sum_{j=1}^{|\mathbb{R}_T|}log(P_w^{(r)}=S_w^{(r)}|t_j,\Theta ) )
\end{equation}

\begin{equation}
\label{eqn:global_objective}
L =  L_{(c)} + \alpha * L_{(r)}
\end{equation}
where $|\mathbb{R}_T|$ is the size of the training dataset, $t_i$ and $t_j$ is the $i$th and $j$th tweet in the training dataset, respectively. $|W_i|$ is the sentence length. $\alpha \in [0,1]$ is a 
trade-off coefficient between loss of concern and concern relation, and the larger value means the greater influence of concern relation on the proposed method.

\section{Experiments}
In this section, extensive experiments are conducted to evaluate the proposed approach by using COVID-19 Twitter datasets. First, COVID-19 dataset collection and pre-processing are described. Second, we compare the proposed approach against six state-of-the-art baselines in terms of accuracy, recall, and F1 score. Third, we present quantitative analytical results and conduct ablation studies. Finally, a case study is given to illustrate the effectiveness of our approach.

\subsection{Dataset and Experiment Setting}
The experiments are conducted by using a public large-scale Twitter dataset about COVID-19, which contains English language-specific tweets being posted from 204 different countries and territories (\cite{lamsal2020design}). The dataset is proposed in the scientific literature for research with topics related to COVID-19. The statistics of datasets are listed in Table \ref{table:data_train_test}. 

\begin{table}
\centering
\caption{Statistics of manual-labelled and auto-labelled dataset}
\begin{tabular}{l|ll} 
\hline
               & Train & Test  \\ 
\hline
Manual-labelled & 1418  & 355   \\ 
\hline
Auto-labelled   & 32264 & 8066  \\
\hline
\end{tabular}

\label{table:data_train_test}
\end{table}

The dataset has been pre-processed in two ways, i.e., manual annotated and auto-annotated. In the former, the annotators label the tweets according to the concern definitions and formulations. While, in the latter, tweets are annotated by using the approach proposed in our past research work (\cite{shi2020concern}). 

Many existing research works have explored people's reactions and attempt to discovered wide-spreading topics about COVID-19 (\cite{li2020characterizing, killeen2020county,hou2020assessment, kaveh2020track, li2020characterizing}). Based on the findings and conclusion of these works, we extract the most popular topics and define eight types of concerns, i.e., Finance (FIN), Government (GOV), Disease (DIS), Medicine (MED), Person (PER), Location (LOC), Food (FOD), and Date and Time (DAT). On top of that, two types of relations among the concerns, i.e., co-occurrence and cause-effect, are investigated. This is because both types of relations are capable of capturing the implicit information about public concerns, demonstrating their associations and potential causes. For instance, by analysing the tweet ``... due to the locked transportation..., farmers forced to dump green chilli ...", it is important to know the concern ``green chilli" is dumped due to the concern ``locked transportation" in the time of COVID-19 pandemic. The statistics of concerns and the relations are listed in Table \ref{table:tweet_type} and Table \ref{table:tweet_relation}, respectively.

The dataset is divided into 3 sub-datasets: train dataset and test dataset, occupying 80\% and 20\%, respectively. 

\begin{table}
\centering
\caption{Statistics of the Concern Categories}
\resizebox{\textwidth}{!}{
\begin{tabular}{clllllllll}
\hline
\multirow{2}{*}{Type} & \multicolumn{1}{c}{\multirow{2}{*}{Tweets}} & \multicolumn{8}{c}{Concern Category}                      \\ \cline{3-10} 
                      & \multicolumn{1}{c}{}                        & FIN  & GOV   & DIS   & MED  & PER   & LOC  & FOD  & DAT   \\ \hline
Manual-labelled        & 1761                                        & 315  & 457   & 1239  & 471  & 289   & 341  & 204  & 206   \\ \hline
Auto-labelled          & 40068                                       & 4341 & 19941 & 23853 & 6944 & 10977 & 1519 & 1498 & 11063 \\ \hline
\end{tabular}
}
\label{table:tweet_type}
\end{table}

\begin{table}[!ht]
\small
\centering
\caption{Statistics of Concern Relation Categories}

\begin{tabular}{clll}
\hline
\multirow{2}{*}{Type} & \multicolumn{1}{c}{\multirow{2}{*}{Tweets}} & \multicolumn{2}{l}{Concern Relation} \\ \cline{3-4} 
                      & \multicolumn{1}{c}{}                        & CO\_OCC           & CA\_EFF          \\ \hline
Manual-labelled        & 1761                                        & 932               & 829              \\ \hline
Auto-labelled          & 40068                                       & 19485             & 20583            \\ \hline
\end{tabular}

\label{table:tweet_relation}
\end{table}

\subsubsection{Evaluation Metrics}
In this paper, three standard evaluation metrics, i.e., precision, recall and F1 score, are employed to evaluate our model. 

The outcome of predicted concerns is considered correct only when both of the concerns in one tweet are predicted correctly. In other words, $(c1, c2)$ is recognised as a correct concern pair if $c1$ and $c2$ are correctly predicted at same time. Correspondingly, the relation prediction is considered as valid only when the associated concern pair is correctly predicted.

\subsubsection{Hyper-parameters}
In the experiments, BERT is utilised to obtain word embedding of tweet corpus and the word embeddings dimension is set as $d=300$. Our network is regularised by using dropout at the embedding layer, with a dropout ratio of 0.2. Bi-LSTM and GCN are adopted as the encoding layer, with 300 LSTM units. We employ the full dependency tree of sentences as the adjacency matrix of GCN.

\subsection{Baselines}  
The proposed approach is evaluated by comparing against the following baselines.

\begin{itemize}
	\item \textbf{\emph{Joint Model }}(\cite{zheng2017joint} is a joint extraction method to detect both entity and relation in one tweet by using a novel tagging scheme. It is an end-to-end model consisting of a bi-directional Long Short Term Memory (Bi-LSTM) encoder layer and a Long Short Term Memory (LSTM) decoder layer.
	\item \textbf{\emph{Copy Mechanism Model }}(\cite{zeng2018extracting} is a state-of-the-art model for jointly extracting relation triplets from a sentence. It is also an end-to-end model based on seq-to-seq learning with a decoder layer, having two different decoding methods, i.e., one-decoder and multi-decoder. We use both different strategies as counterparts in the experiments.
	\item \textbf{\emph{SPTree}}(\cite{miwa2016end} is a novel end-to-end recurrent neural network model, aiming at extracting entities and relations by capturing word sequence and dependency tree substructure feature. The stacked bidirectional tree-structured LSTM-RNN models are applied on sequential bi-LSTM-RNN models to detect both entities and relations with shared parameters jointly.
	\item \textbf{\emph{JointER}}(\cite{yu2020jointer} is a joint entity and relation extraction model which can address the limitations, including redundant entity pairs and ignoring the important inner structure of entities. The model decomposes a joint extraction task into Head-Entity (HE) extraction and Tail-Entity-Relation (TER) extraction to detect head-entity, tail-entity, and relations.
	\item \textbf{\emph{SPERT}}(\cite{eberts2019span} is introduced as a span-based model, which can jointly extract entity and relation by conducting light-weight reasoning on BERT embedding and relation classification based on localised and marker-free context features. 
\end{itemize}

\subsection{Experimental Results and Model Analysis}

In this section, we present and analyse the strengths and weaknesses of the proposed method by comparing against the state-of-the-art models as mentioned previously. The ensure the fairness and rationality of the experiments, we select all the counterparts, which incorporate a Bi-LSTM encoder layer. 

\begin{figure}
	\centering
	\includegraphics[width=0.8\textwidth]{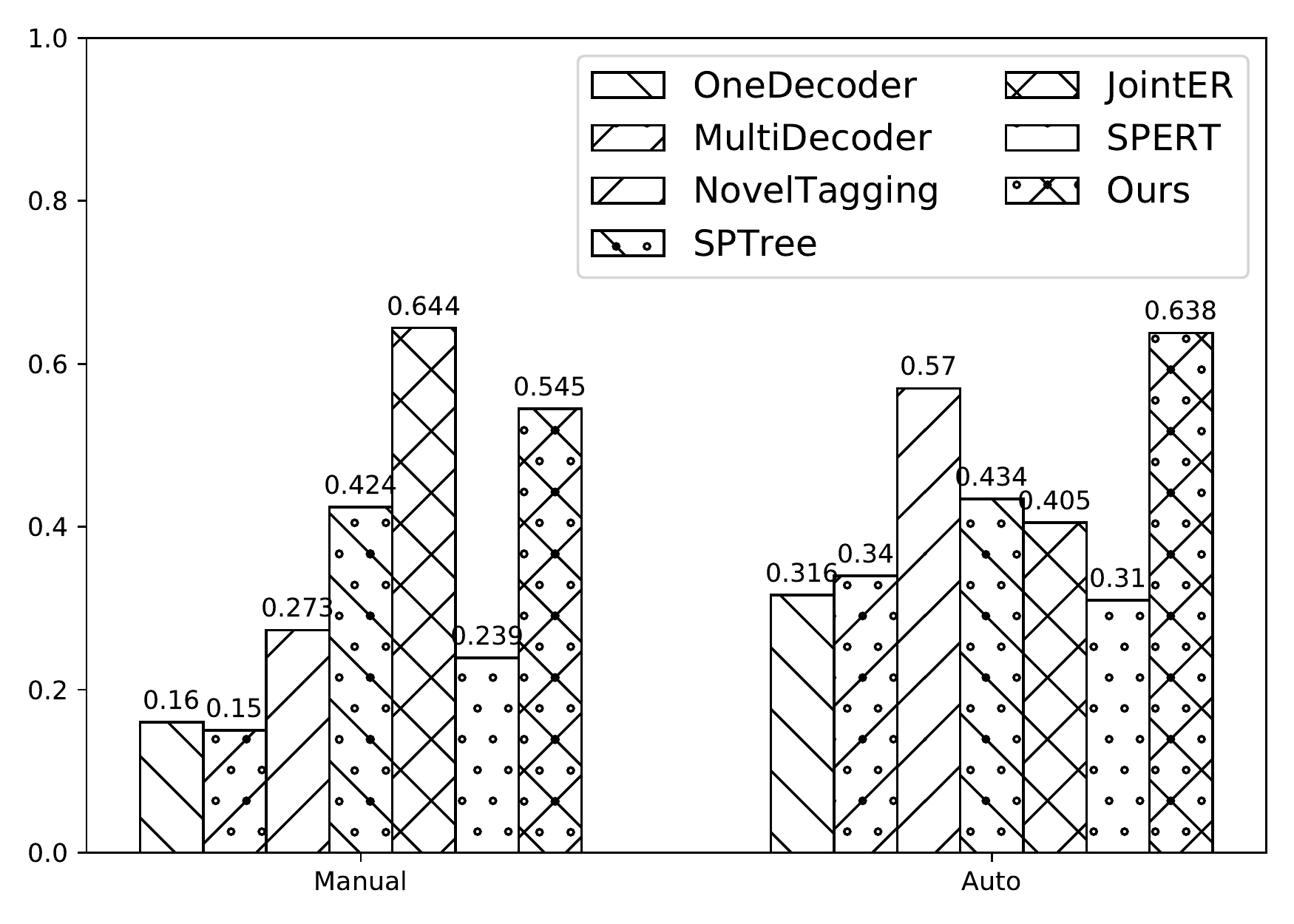}
	\caption{Experiment results (Precision) on manual-labelled dataset.}
	\label{fig:res_precision}    
\end{figure}

\begin{figure}
	\centering
	\includegraphics[width=0.8\textwidth]{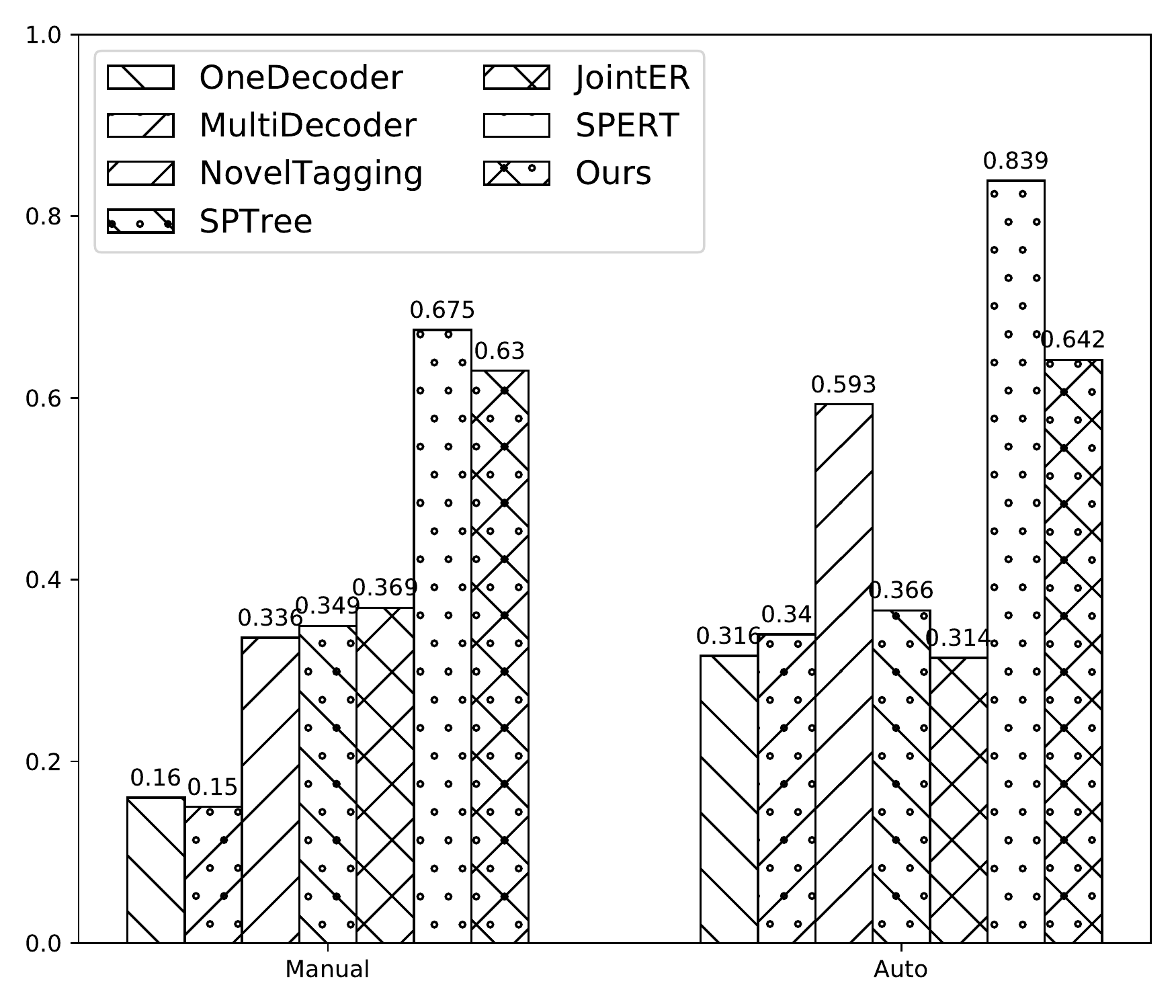}
	\caption{Experiment results (Recall) on manual-labelled dataset.}
	\label{fig:res_recall}       
\end{figure}

\begin{figure}
	\centering
	\includegraphics[width=0.8\textwidth]{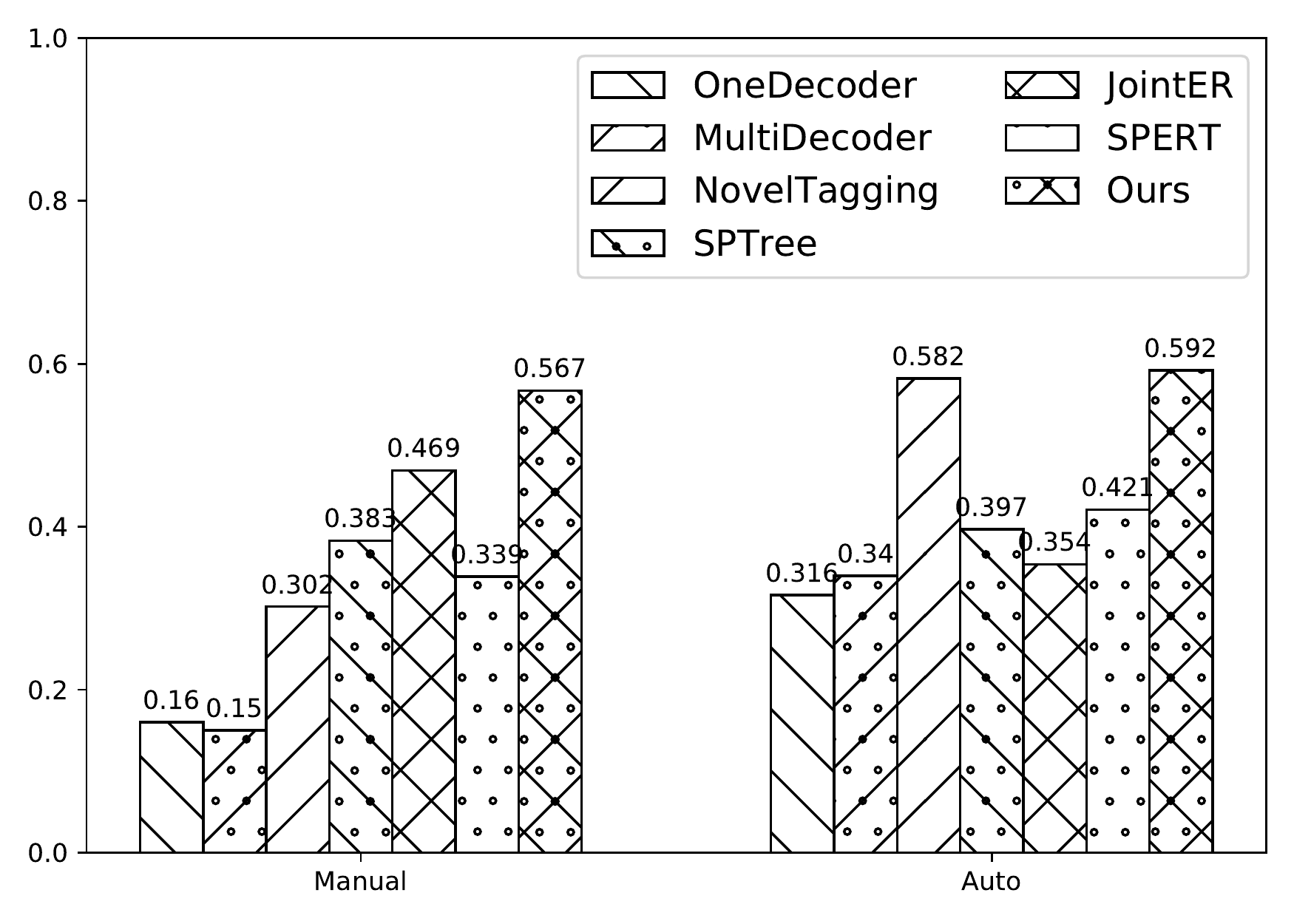}
	\caption{Experiment results (F1 score) on manual-labelled dataset..}
	\label{fig:res_f1}   
\end{figure}

The experimental results are demonstrated in Table \ref{table:predicted_results}, which presents the predicted outcome, i.e., Precision, Recall, and F1, of both the proposed approach and the state-of-the-art methods, using  manual-labelled and auto-labelled datasets. As can be observed from the table that the proposed approach outperforms the others in terms of F1 score, which proves its effectiveness. Specifically, in Figures \ref{fig:res_precision} - \ref{fig:res_f1}, the CG-based model outperforms One-decoder, Multi-decoder, NovelTagging, and SPTree models on both manual-labelled and auto-labelled Twitter datasets. Although JointER and SPERT yield better performance than that of ours in terms of precision and recall on the manual-labelled dataset, SPERT leverages the pre-trained BERT model to obtain contextual features of sentences but the inner structure of entities is neglected, which inevitably hinders the performance of entity and relation extraction. The embeddings in JointER are initialized using  the shallow representatives model, i.e., Glove (\cite{pennington2014glove},  without context-specific information, which is critical for entity and relation extraction models.

The outstanding performance of the proposed approach mainly attributes to its structural design. First, the interaction of the CG structure captures the inner dependency between concerns. Second, the shared state passing from concern extraction module to relation extraction module, provides important concern features for relation extraction. It is worth noting that baselines can achieve state-of-the-art results on high-quality datasets, e.g., NYT and WebNLG, but the performance significantly degrades on the noisy and imbalanced social media data. The grammatical mistakes of tweets make it difficult to capture relations between concerns. NovelTagging and SPTree utilise novel tagging, but cannot carry out promising results. Other baselines, including One-Decoder, Multi-Decoder, JointER, apply BiLSTM to capture sequential features of concerns but they fail to detect the relation features and concerns due to the unstructured sentences in the tweet dataset.

\begin{table}[h]
	\centering
	\caption{Evaluation results of different models on COVID-19 Tweets Dataset}
	\resizebox{\textwidth}{!}{
		\begin{tabular}{l|lll|lll}
			\hline
			\multicolumn{1}{c|}{\multirow{2}{*}{Model}} & \multicolumn{3}{l|}{Manual-labelled Tweets} & \multicolumn{3}{l}{Auto-labelled Tweets} \\ \cline{2-7} 
			\multicolumn{1}{c|}{}                       & Precision        & Recall       & F1       & Precision      & Recall      & F1        \\ \hline
			One-Decoder                                 &0.160                  &0.160              &0.160          &0.316                &0.316             & 0.316     \\ \hline
			Multi-Decoder                               &  0.150                & 0.150             &  0.150        & 0.340               &0.340             & 0.340     \\ \hline
			NovelTagging                                &   0.273               &   0.336           & 0.302         &0.570               &0.593             & 0.582     \\ \hline
			SPTree                                &   0.424              &   0.349           & 0.383         &0.434               &0.366            & 0.397     \\ \hline
			JointER                                &\textbf{0.644}              &   0.369           & 0.469         & 0.405              &0.314           &0.354     \\ \hline
			SPERT                                &  0.239              &\textbf{0.675}            &0.339         &  0.31             &\textbf{0.839}           & 0.421    \\ \hline
			
			Proposed Model                              &  0.545               &     0.630         &     \textbf{0.567 }    &    \textbf{0.638}           &  0.642          &    \textbf{0.592   }    \\  \hline
		\end{tabular}
	}
	\label{table:predicted_results}
\end{table}

\subsection{Ablation Study}
The ablation study in this section aims to investigate the impact of Concern Graph and shared state components in the proposed approach. 

Recall that manually labelling a large-scale dataset turns out to be a tedious and non-trivial task. To conduct public concern extraction and analysis for an emergency event, sufficient manual-labelled training data sets are usually not available. Furthermore, the public concern coverage in datasets also appears imbalanced, which prevents the existing models from generalisation, subsequently impacting the performance to a large extent. The proposed approach can mitigate this issue, giving outstanding performance on both manual-labelled and auto-labelled datasets. 

Table \ref{table:contribution_analysis} lists the results of the ablation study. The approach has been re-evaluated by comparing the performance against that without CG component and shared state components. It can be seen from the table that, in manual-labelled dataset, CG-CRE with CG and shared state outperforms the models without CG and shared state by 11\% and 7\%, respectively. While in auto-labelled datasets, it surpasses 6\% and 1\%, respectively. The results explicitly reveal that Concern Graph (CG) and shared state components play a significant role in jointly identifying concerns and relations.

\begin{table}
	\centering
	\caption{Ablation study of CG-CRE model on manual-labelled and auto-labelled Tweets Dataset}
	\resizebox{\textwidth}{!}{
		\begin{tabular}{cllll}
			\hline
			Dataset                                & \multicolumn{1}{c}{Method}    & \multicolumn{1}{c}{Precision} & Recall & F1 \\ \hline
			\multirow{3}{*}{Manual-labelled Tweets} & CG-CRE (without CG)           & 0.416                         & 0.482      &  0.457  \\
			& CG-CRE (without shared state) & 0.463                          &0.516       &   0.494 \\
			& CG-CRE (with all components)  & \textbf{0.545 }                             & \textbf{0.630}       & \textbf{0.567}   \\ \hline
			\multirow{3}{*}{Auto-labelled Tweets}   & CG-CRE (without CG)           & 0.551                        & 0.583       &0.536    \\
			& CG-CRE (without shared state) &       0.615                        &   0.624    &  0.586  \\
			& CG-CRE (with all components)  & \textbf{0.638}                          &     \textbf{0.642}   &  \textbf{0.592}  \\ \hline
		\end{tabular}
	}
	\label{table:contribution_analysis}
\end{table}

\subsection{Case Study}
In this section, we conduct case studies, presenting some representative public concern extraction examples, to further prove the effectiveness and validity of the proposed approach. 

Table \ref{table:case_study} shows the outputs from three models, including NovelTagging, JointER, and the proposed CG-CRE. In the first case, both concerns and concern relation are identified incorrectly by NovelTagging, and JointER predicts nothing. By contrast, CG-CRE can extract both concerns correctly. Similar outputs is presented in the fifth and the sixth case. As for the second and third case, NovelTagging only detects one concern correctly and cannot extract the second one and relation. However, JointER and CG-CRE can accurately identify concerns and concern relation. JointER is not able to carry out the prediction results. In the fourth case, NovelTagging can identify only one concern correctly. JointER is able to yield correct predictions but it still remains to be improved in eliminating null prediction. NovelTagging is weak at extracting relations from Twitter datasets. 

Based on the experimental results and case studies, we can conclude that the proposed CG-CRE model can yield better performance on both entity recognition and relation extraction than the state-of-the-art models.

\begin{table}[]
\centering
\caption{Outputs from different models on tweets. ``pred:[]'' means the model predicts null for this tweet. NovelTagging only predicts ``c1" and ``c2" without concern types.}
\resizebox{\textwidth}{!}{
\begin{tabular}{cl}
\hline
Models       & \multicolumn{1}{c}{Tweet}                                                                                        \\ \hline
NovelTagging & \begin{tabular}[c]{@{}l@{}} $[\textbf{seeing everyone}]_{{c1}, {r:co\_{occ}}}$ suffer $[\textbf{in pakistan}]_{{c2}, {r:co\_{occ}}}$ with food shortage \\ due to the corona virus i have made bag which contain rice\end{tabular} \\
JointER      & \begin{tabular}[c]{@{}l@{}}seeing everyone suffer in pakistan with food shortage \\  due to the corona virus i have made bag which contain rice $[\textbf{pred:[]}]$\end{tabular} \\
CG-CRE       & \begin{tabular}[c]{@{}l@{}}seeing everyone suffer in pakistan with $[\textbf{food shortage}]_{{c1:FOD}, {r:ca\_{eff}}}$ \\  due to the $[\textbf{corona virus}]_{{c2:DIS}, {r:ca\_{eff}}}$ i have made bag which contain rice\end{tabular} \\ 

NovelTagging & \begin{tabular}[c]{@{}l@{}}a greeting from the heart to $[\textbf{doctors}]_{{c1}, {r:co\_{occ}}}$ , nurses, $[\textbf{paramedics}]_{{c2}, {r:co\_{occ}}}$, ... \\who stand together to tackle the corona epidemic.\end{tabular} \\
JointER      & \begin{tabular}[c]{@{}l@{}}a greeting from the heart to $[\textbf{doctors}]_{{c1:MED}, {r:co\_{occ}}}$ , $[\textbf{nurses}]_{{c2:MED}, {r:co\_{occ}}}$, paramedics, ... \\who stand together to tackle the corona epidemic.\end{tabular} \\
CG-CRE       & \begin{tabular}[c]{@{}l@{}}a greeting from the heart to  $[\textbf{doctors}]_{{c1:MED}, {r:co\_{occ}}}$ ,$[\textbf{nurses}]_{{c2:MED}, {r:co\_{occ}}}$, paramedics, ... \\who stand together to tackle the corona epidemic.\end{tabular} \\

NovelTagging & \begin{tabular}[c]{@{}l@{}}$[\textbf{coronavirus}]_{{c1}, {r:co\_{occ}}}$ could double number of people going hungry. \\ the risk of major interruptions to $[\textbf{food supplies}]_{{c21}, {r:co\_{occ}}}$ over the coming months is growing.\end{tabular} \\
JointER      & \begin{tabular}[c]{@{}l@{}}$[\textbf{coronavirus}]_{{c1:DIS}, {r:ca\_{eff}}}$ could double number of people $[\textbf{going hungry}]_{{c2:FOD}, {r:ca\_{eff}}}$. \\ the risk of major interruptions to food supplies over the coming months is growing.\end{tabular} \\
CG-CRE       & \begin{tabular}[c]{@{}l@{}}$[\textbf{coronavirus}]_{{c1:DIS}, {r:ca\_{eff}}}$ could double number of people $[\textbf{going hungry}]_{{c2:FOD}, {r:ca\_{eff}}}$. \\ the risk of major interruptions to food supplies over the coming months is growing.\end{tabular} \\

NovelTagging & \begin{tabular}[c]{@{}l@{}}breaking one of somalia ’s greatest artist ha $[\textbf{died}]_{{c1}, {r:co\_{occ}}}$ in london \\ after contracting $[\textbf{corona virus}]_{{c2}, {r:co\_{occ}}}$ ...\end{tabular} \\
JointER      & \begin{tabular}[c]{@{}l@{}}breaking one of somalia ’s greatest artist ha died in london \\ after contracting corona virus ... $[\textbf{pred:[]}]$\end{tabular} \\
CG-CRE       & \begin{tabular}[c]{@{}l@{}}breaking one of somalia ’s greatest $[\textbf{artist}]_{{c1:PER}, {r:ca\_{eff}}}$ ha died in london \\ after contracting $[\textbf{corona virus}]_{{c2:DIS}, {r:ca\_{eff}}}$ ...\end{tabular} \\

NovelTagging & \begin{tabular}[c]{@{}l@{}}what are the $[\textbf{common}]_{{c1}, {r:co\_{occ}}}$ $[\textbf{symptom}]_{{c2}, {r:co\_{occ}}}$ . fever , sore throat ...\end{tabular} \\
JointER      & \begin{tabular}[c]{@{}l@{}}what are the common symptom . fever , sore throat ... $[\textbf{pred:[]}]$\end{tabular} \\
CG-CRE       & \begin{tabular}[c]{@{}l@{}}what are the common symptom . $[\textbf{fever}]_{{c1:DIS}, {r:co\_{occ}}}$ , $[\textbf{sore throat}]_{{c2:DIS}, {r:co\_{occ}}}$ ...\end{tabular} \\

NovelTagging & \begin{tabular}[c]{@{}l@{}}social distancing, stay home, $[\textbf{naija people}]_{{c1}, {r:co\_{occ}}}$  will not hear. \\ this corona thing has just started with us in this $[\textbf{country}]_{{c2}, {r:co\_{occ}}}$, we ...\end{tabular} \\
JointER      & \begin{tabular}[c]{@{}l@{}}social distancing, stay home, naija people will not hear. \\ this corona thing has just started with us in this country, we ... $[\textbf{pred:[]}]$\end{tabular} \\
CG-CRE       & \begin{tabular}[c]{@{}l@{}}$[\textbf{social distancing}]_{{c1:GOV}, {r:co\_{occ}}}$, $[\textbf{stay home}]_{{c2:GOV}, {r:co\_{occ}}}$, naija people will not hear. \\ this corona thing has just started with us in this country, we ...\end{tabular} \\ \hline
\end{tabular}
}
\label{table:case_study}
\end{table}

\section{Conclusion and Future Work}
In this paper, we present an end-to-end model to simultaneously extract concern and concern relation from the social media dataset of COVID-19.  We jointly combine GCN and Bi-LSTM to learn sequential and regional dependency features from tweets. In order to capture more features of model input, the influence of graph structure for concern and relation extraction is explored. The sequential and regional features from the dataset are concatenated, enabling the embedding vectors to represent rich contextual information of both concerns and relations. The proposed model is evaluated on manual-labelled and auto-labelled datasets. The experimental results show that the proposed model can outperform the existing entity and relation extraction models, which demonstrate the effectiveness of our method. Furthermore, different from previous works, our model turns out to be applicable for both manual-labelled and auto-labelled datasets rather than only works with handcrafted datasets. Therefore, our method can be easily transferred and applied to other pandemics situations, e.g., Zika, Dengue Fever,  and Yellow Fever.

In the future, we plan to improve the proposed model from two aspects. First, more concern types and concern relation types can be predicted to understand what people's attention and the relation between them. In addition, time factor can be used to track the trend of one specific concern over time.

\bibliographystyle{unsrtnat}
\bibliography{paper2_arxiv}  

\end{document}